\pdfoutput=1

\documentclass[11pt]{article}

\usepackage[]{acl}

\usepackage{times}
\usepackage{latexsym}
\usepackage{booktabs}
\usepackage{graphicx}
\usepackage{adjustbox}
\usepackage{multirow}
\usepackage{subcaption}
\usepackage{amsmath}
\usepackage{amsfonts}
\usepackage{amssymb}

\usepackage[T1]{fontenc}

\usepackage[utf8]{inputenc}

\usepackage{microtype}

\usepackage{inconsolata}

\title{Leveraging Open Information Extraction for \\ More Robust Domain Transfer of Event Trigger Detection}

\author{David Dukić$^{1,\dagger}$ \and Kiril Gashteovski$^{2,3}$ \and Goran Glavaš$^{4}$  \and Jan Šnajder$^1$  \\ $^1$TakeLab, Faculty of Electrical Engineering and Computing, University of Zagreb \\ $^2$NEC Laboratories Europe, Heidelberg, Germany \\ $^3$CAIR, Ss.~Cyril and Methodius University, Skopje, North Macedonia \\ 
$^4$CAIDAS, University of Würzburg, Germany \\ $^{1,2,4}$\texttt{name.surname@\{fer.hr, neclab.eu, uni-wuerzburg.de\}}}

\begin{document}
\maketitle
\def\thefootnote{$\dagger$}\footnotetext{Corresponding author: \texttt{david.dukic@fer.hr}}
\renewcommand{\thefootnote}{\arabic{footnote}}
\begin{abstract}
Event detection is a crucial information extraction task in many domains, such as Wikipedia or news. The task typically relies on trigger detection (TD) -- identifying token spans in the text that evoke specific events. While the notion of triggers should ideally be universal across domains, domain transfer for TD from high- to low-resource domains results in significant performance drops. We address the problem of negative transfer in TD by coupling triggers between domains using subject-object relations obtained from a rule-based open information extraction (OIE) system. We demonstrate that OIE relations injected through multi-task training can act as mediators between triggers in different domains, enhancing zero- and few-shot TD domain transfer and reducing performance drops, in particular when transferring from a high-resource source domain (Wikipedia) to a low(er)-resource target domain (news). Additionally, we combine this improved transfer with masked language modeling on the target domain, observing further TD transfer gains. Finally, we demonstrate that the gains are robust to the choice of the OIE system.\footnote{\footnotesize{Find code at \url{https://github.com/dd1497/oie-td}.}}
\end{abstract}

\section{Introduction}

Event detection is an important part of the information extraction pipeline in natural language processing (NLP). Event detection systems are typically bound to domain-specific schemes and fill predefined event-specific slots evoked by an event \emph{trigger} -- a span of words that 
evokes a particular type of event. A typical domain-specific event detection workflow consists of trigger detection (TD), which locates the trigger span in the text, and trigger classification \citep{xiang2019survey}, which assigns one of the predefined event types to the trigger. With triggers identified, the next step is typically to detect the corresponding arguments, e.g., participants, location, and time. 
The detected events can be leveraged for many downstream tasks, including knowledge graph construction \citep{zhang-etal-2021-eventke-event}, information retrieval \citep{glavas-snajder-2013-event}, text summarization \citep{zhang-etal-2023-enhancing}, and aspect-based sentiment analysis \citep{tang-etal-2022-affective}.

\begin{figure}
    \centering
    \includegraphics[width=\columnwidth]{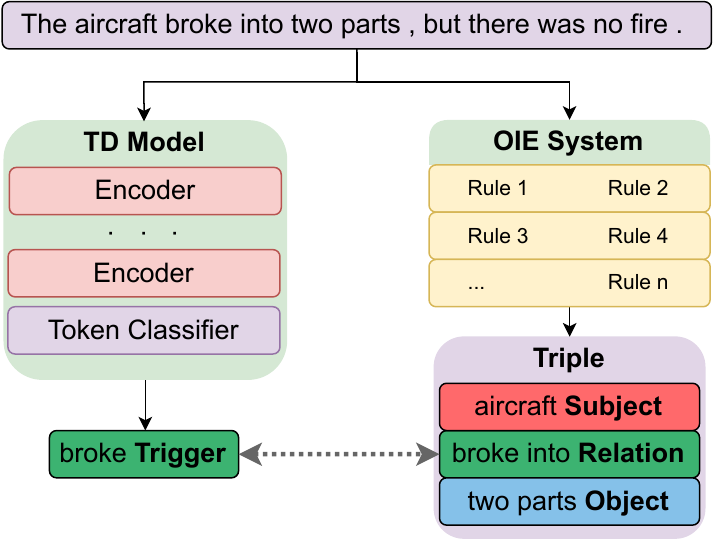}
    \caption{An example of event trigger detection and subject-relation-object extraction with an open information extraction (OIE) system. The detected trigger and extracted OIE relation often overlap to a significant degree, which can be leveraged for creating more robust trigger detection models across domains.
    }
    \label{fig:triple_extraction}
\end{figure}

While the notion of an event trigger is intuitive and universal (i.e., events and their triggers exist in all text domains), 
NLP research has struggled to provide a clear-cut operational definition of an event, giving rise to diverse annotation schemes, e.g., 
\cite{doddington-etal-2004-automatic,
pustejovsky2005specification,shaw2009lode,cybulska2014guidelines,song-etal-2015-light}. The differences between annotation schemes, alongside the usual distribution shifts between text domains, make domain transfer of TD very challenging.  
Empirical evidence has demonstrated massive performance drops in zero- and low few-shot TD transfer from a high-resource source to a low(er)-resource target domain -- a phenomenon commonly referred to as \emph{negative transfer} \citep{wang2019characterizing, ngo-trung-etal-2021-unsupervised, meftah-etal-2021-hidden}. 
The absence of an effective domain transfer method for TD implies a costly (large-scale) manual annotation of event trigger spans for each domain of interest. 



One way to facilitate domain transfer of TD may be by means of a proxy task that (i) exhibits a smaller distributional shift across domains and could thus (ii) mediate representational alignment between triggers of different domains.  
In principle, all tasks that extract structures that relate to event semantics, such as syntactic or predicate-argument structures, make good candidates for such a mediator \citep{mcclosky-etal-2011-event, liu-etal-2016-leveraging}. 
Recent work by \citet{deng-etal-2022-title2event} showed that trigger and argument detection could be aligned with the subject-relation-object triples as mediators (in Chinese), with subjects and objects mapped to arguments and relations to triggers. In other words, both events and subject-relation-object triples represent predicate-argument structures, pointing to tasks that extract the latter as potentially good mediators for domain transfer of TD.  


Open Information Extraction (OIE) systems \citep{banko2007open} automatically extract subject-relation-object triples in a domain-independent manner because they discover relations not predefined by any schema \citep{fader-etal-2011-identifying, biomedoie, sun-etal-2018-logician, gashteovski2019opiec}. Although most recent OIE systems are neural models trained in a supervised manner \citep{kolluru-etal-2020-openie6,kotnis-etal-2022-milie}, traditional OIE systems such as Stanford OIE \citep{angeli-etal-2015-leveraging} and MinIE \citep{gashteovski-etal-2017-minie} are rule-based and typically do not require domain-specific pre-processing of the input text \citep{lauscher2019minscie}. Moreover, recent fact-based evaluation \cite{gashteovski-etal-2022-benchie} renders them more accurate than neural OIE models.   
Figure~\ref{fig:triple_extraction} illustrates the overlap between the trigger \emph{broke} detected by the trigger detection model and an OIE relation \emph{broke into}, extracted by MinIE. 
This overlap is the main motivation for our work. 

In this paper, we address the challenge of negative transfer in TD by leveraging OIE relations to align representations of event triggers across domains. While annotating event triggers in the target domain is costly, automatic extraction of open relations with a rule-based OIE system is cheap, even at a large scale. 
With this in mind, we investigate remedies for negative domain transfer of TD based on the automatic extraction of OIE subject-object relations.
More precisely, we couple the domain-specific trigger annotations with the relation extractions obtained with a domain-agnostic rule-based OIE system through different (i) multi-task architectures and (ii) zero- and few-shot transfer regimes. The intuition is that, by coupling trigger annotations with OIE relations, we effectively couple event triggers between domains with OIE relations as mediators.
Although OIE relations do not always align perfectly with event triggers, we find that they can facilitate and stabilize the domain transfer of TD. 
We demonstrate that (i) multi-task fine-tuning of a pretrained language model (PLM) for OIE relation extraction and TD   and (ii) transfer training regimes adopted from the body of work on language transfer \citep{lauscher-etal-2020-zero,schmidt-etal-2022-dont} reduce the trigger distribution shift between domains and consequently improve TD performance in the low-resource target domain.


\noindent\textbf{Contributions.} \textbf{(1)} We mitigate negative domain transfer of trigger detection by coupling event triggers with subject-object relations extracted by rule-based OIE; we couple the two in different multi-task model designs and investigate the effects in both zero- and few-shot transfer. \textbf{(2)} We show that target-domain masked language modeling (MLM), in the vein of \citet{gururangan-etal-2020-dont}, as an additional auxiliary objective next to open relation extraction, further improves TD transfer. \textbf{(3)} We validate that the gains from the OIE-based proxy are robust and not dependent on the specific OIE system. We believe our work is an important step towards universally more effective event extraction. 




\section{Background and Related Work}

\paragraph{Domain Transfer.} Domain transfer has been investigated for numerous structured prediction tasks such as query translation \citep{yao-etal-2020-domain}, term extraction \citep{hazem-etal-2022-cross}, named entity recognition \citep{jia-zhang-2020-multi} and disambiguation \citep{blair-bar-2022-improving}, and event argument extraction \cite{sainz-etal-2022-textual}. 
Existing work on domain transfer for event extraction predominantly resorted to   
semantic role labeling (SRL) as the vehicle for facilitating the transfer. \citet{lyu-etal-2021-zero} ran SRL to detect predicates as potential event triggers for the domain transfer of event extraction via question answering and textual entailment models. \citet{peng-etal-2016-event} investigated the use of SRL predicates and arguments to facilitate domain transfer for both event detection and event co-reference resolution. While SRL is structurally fit to be a proxy task for even extraction, it is also a task that requires domain-specific annotations. More recently, domain adaptation for models based on PLMs has been driven by general self-supervised language modeling on (unlabeled) domain-specific corpora \cite{gururangan-etal-2020-dont,hung-etal-2022-ds}.   

\paragraph{Domain Adaptation for Event Detection.} \citet{nguyen-grishman-2015-event} were the first to employ a convolutional neural network (CNN) for event detection domain adaptation by learning more universal trigger representations through a CNN architecture and various features such as word, position, and entity type embeddings. \citet{naik-rose-2020-towards} tackled TD transfer between literature and news domains using adversarial domain adaptation to produce representations predictive for triggers but not predictive of the example's domain, thus forcing the model to learn domain-agnostic trigger representations. \citet{ngo-trung-etal-2021-unsupervised} leveraged domain-specific adapters for event detection domain transfer. More recently, \citet{trung-etal-2022-unsupervised} developed an unsupervised domain adaptation method applicable to text classification tasks, including event detection and sentiment classification, which utilizes meta- and self-paced learning approaches. 
Other strands of research deal with improving few-shot event detection but are mostly limited to in-domain transfer between different event types \citep{lai-etal-2020-extensively, li-etal-2020-event}. Examples include improving the zero- and few-shot in-domain event detection performance with cloze-based prompt meta-learning \citep{yue2023zero} and ontology embeddings \citep{deng-etal-2021-ontoed}.


\paragraph{OIE for NLP tasks.} OIE systems are intended to facilitate various downstream tasks, including text summarization \cite{fan-etal-2019-using,ribeiro-etal-2022-factgraph}, question answering \cite{yan2018assertion, nagumothu-etal-2022-pie}, incomplete sentence reconstruction \citep{montella-etal-2020-denoising}, and event extraction \citep{chen-etal-2023-led}. Many event-related tasks, such as event schema induction \citep{balasubramanian-etal-2013-generating} and cross-domain event coreference \citep{pratapa-etal-2021-cross}, benefit from leveraging OIE triples. However, OIE has not yet been employed to improve TD. A step in that direction is the work by \citet{deng-etal-2022-title2event}, where authors created a dataset named \emph{Title2Event} consisting of Chinese titles designed for \emph{open event extraction} based on OIE triples, subscribing to the idea that events are well-aligned with the subject-relation-object schema, which we also adopt in this work.



\section{OIE for Event Trigger Detection} \label{sec:oie_for_td}

Following prior work \citep{naik-rose-2020-towards, ngo-trung-etal-2021-unsupervised}, we frame TD as a sequence labeling task where each token is classified as either part of some event trigger span or outside of it. This task formulation is intuitive, given that event triggers are consecutive token sequences, and multiple triggers may appear in the same input sentence.  
We use the widely adopted
IOB2 (inside, outside, begin) tagging scheme \citep{iob2}. 
Analogously, we model relation extraction (RE) -- for which we use OIE relation extractions as ground-truth labels -- also as a sequence labeling task with its own set of IOB2 tags.
We tackle domain transfer for TD with two different model architectures (based on a PLM) that couple OIE relations with TD annotations, which we refer to as (i) \emph{implicit} and (ii) \emph{explicit} OIE-TD multi-task models. We next describe both variants in detail.   



\begin{figure}
    \centering
    \includegraphics[width=0.9\columnwidth]{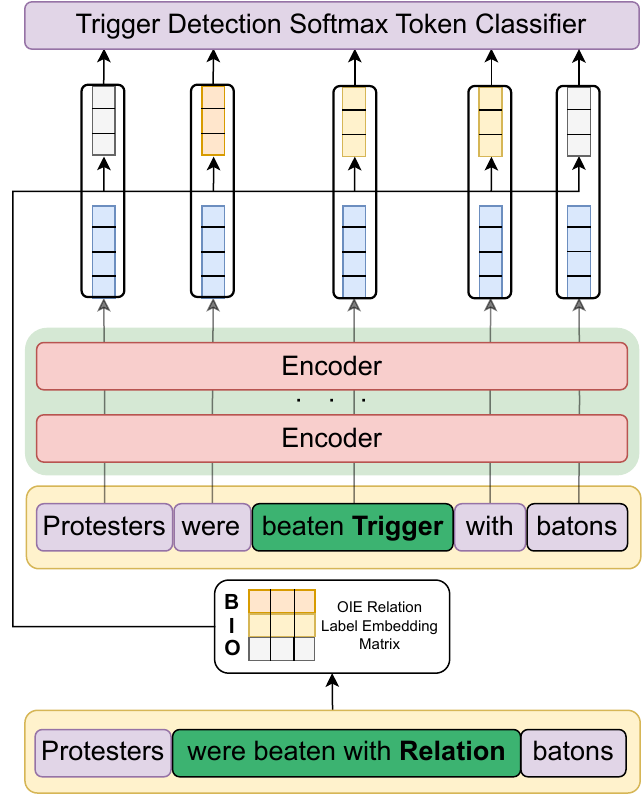}
    \caption{\emph{Implicit} model during training. The input sentence is fed twice: once with trigger IOB2 tags through PLM encoders and once with OIE relation IOB2 tags by indexing the corresponding label embedding matrix. At the \emph{implicit} output, PLM's last hidden state embeddings are concatenated with OIE relation label embeddings per token and passed through the TD softmax classifier.}
    \label{fig:implicit_model}
\end{figure}

\paragraph{Implicit Multi-Task.}

In the \emph{implicit} model, we train and use embeddings for token labels of OIE relations: one randomly initialized vector for each of the three IOB2 tags. The model concatenates the embedding $\mathbf{x}_\mathrm{OIE} \in \mathbb{R}^d$ of the OIE relation label of each token embedding to the contextualized token embedding of the token $\mathbf{x}_\mathrm{PLM} \in \mathbb{R}^h$ (the output of the last PLM layer), where $d$ is the dimension of the trainable OIE relation label embeddings (hyperparameter of the model), and $h$ is the PLM's hidden size. The final token representation, $\mathbf{x} = [\mathbf{x}_\mathrm{PLM}; \mathbf{x}_\mathrm{OIE}]$, is fed to the standard softmax classifier, which predicts the IOB2 event trigger label for the token, $\mathit{softmax}(\mathbf{W}_\mathrm{cl}^\mathrm{T}\mathbf{x} + \mathbf{b}_\mathrm{cl})$, with $\mathbf{W}_\mathrm{cl} \in \mathbb{R}^{(d+h)\times 3}$ and $\mathbf{b}_\mathrm{cl} \in \mathbb{R}^{3}$ as trainable parameters of the classifier. As is common in multi-class classification, we tune all parameters by minimizing the (multi-class) cross-entropy loss. The \emph{implicit} model is illustrated in Figure~\ref{fig:implicit_model}. We train the model on TD in the source domain, optimizing (1) all of the PLM's parameters, (2) classifier's parameters $\mathbf{W}_\mathrm{cl}$ and $\mathbf{b}_\mathrm{cl}$, and (3) embedding matrix $\mathbf{X}_\mathrm{OIE} \in \mathbb{R}^{3\times d}$ containing the trainable embeddings of the OIE labels. At inference time in the target domain, we run the OIE system on test sentences to obtain the OIE relation labels for tokens and then perform inference using the \emph{implicit} PLM for TD and embeddings of OIE labels obtained in training.  



We hypothesize that the \emph{implicit} model is incentivized to establish -- within the OIE label embeddings trained via event TD -- contextualized associations between the two tasks. 
Intuitively, this should improve the recall of TD in the target domain as long as the OIE -- which is rule-based and thus more domain agnostic -- is resilient to distribution shifts between domains.  
Similar event detection approaches based on training label embeddings exist \citep{nguyen-grishman-2015-event, liu-etal-2017-exploiting, ji-etal-2019-exploiting}. However, they typically concatenate the label and token embeddings at the encoder's input and rely on encoders shallower than common Transformer-based PLMs.

\paragraph{Explicit Multi-Task.}


The \emph{explicit} model works with two standard softmax classifiers and a shared PLM encoder. The representation of each token $\mathbf{x}_\mathrm{PLM} \in \mathbb{R}^h$, from PLM's last layer, is forwarded to the (i) TD softmax classifier $\mathit{softmax}(\mathbf{W}_\mathrm{td}^\mathrm{T}\mathbf{x}_\mathrm{PLM} + \mathbf{b}_\mathrm{td})$, which predicts the IOB2 event trigger label for the token and (ii) RE softmax classifier $\mathit{softmax}(\mathbf{W}_\mathrm{re}^\mathrm{T}\mathbf{x}_\mathrm{PLM} + \mathbf{b}_\mathrm{re})$, which predicts the IOB2 relation label for the token, with $\mathbf{W}_\mathrm{td},\mathbf{W}_\mathrm{re} \in \mathbb{R}^{h\times3}$ and $\mathbf{b}_\mathrm{td},\mathbf{b}_\mathrm{re} \in \mathbb{R}^3$ as trainable parameters of two classifiers. Based on the predictions, the (multi-class) cross-entropy loss is calculated for each classifier separately on a mini-batch basis. The average of calculated TD and RE losses is used to update PLM's and classifiers' parameters during training. This is where the interaction of knowledge from both tasks occurs. At inference time, we do not use OIE relation labels in any way. The intuition is that if the notion of triggers is universal across domains and the OIE relations are indeed domain-independent, it should be sufficient only to leverage the in-domain trigger-relation connection during training.  Considering that the TD and RE tasks have the same number of corresponding labels, we tried to share the softmax classifier between TD and RE, but that led to worse overall performance. 




\section{Experimental Setup} \label{sec:exp_setup}


Our experiments investigate the transfer from a high-resource source domain to a low-resource target domain, which is the common transfer direction. For facilitating few-shot domain transfer of TD, we employ \emph{joint} and \emph{sequential} transfer training regimes in combination with multi-task models. 

\subsection{Datasets and Preprocessing} 

As a dataset from a high-resource source domain, we use MAVEN, a dataset of Wikipedia articles with sentence-level trigger annotations. In the low-resource target domain, we use datasets from the news domain -- ACE 2005, EDNYT, and the EVEXTRA -- which also have sentence-level trigger annotations. Table~\ref{tab:dataset_stats} summarizes the dataset statistics.

\begin{table}
  \centering
  \adjustbox{width=\columnwidth}{\small{\setlength{\tabcolsep}{1.75pt}\begin{tabular}{@{}lrrrrrrrrr@{}}
    \toprule
    \multicolumn{1}{c}{\multirow{2}{*}{\textbf{Dataset}}} & \multicolumn{3}{c}{\textbf{Train}} & \multicolumn{3}{c}{\textbf{Valid}} & \multicolumn{3}{c}{\textbf{Test}} \\
    \cmidrule(lr){2-4} \cmidrule(lr){5-7} \cmidrule(lr){8-10}
    {} & \multicolumn{1}{c}{\#Sent} & \multicolumn{1}{c}{\#Tr} & \multicolumn{1}{c}{\#Re} & \multicolumn{1}{c}{\#Sent} & \multicolumn{1}{c}{\#Tr} & \multicolumn{1}{c}{\#Re} & \multicolumn{1}{c}{\#Sent} & \multicolumn{1}{c}{\#Tr} & \multicolumn{1}{c}{\#Re} \\
    \midrule
    MAVEN     & 25944  & 24063 & 15590  & 6487 & 6038 & 3940  & 8042 & 7469 & 4805    \\
    ACE 2005  & 14672 & 3256 & 7403 & 873 & 340 & 446 & 711 & 292 & 412     \\
    EDNYT     & 1842  & 1500 & 1164 & 95 & 74 & 65 & 198 & 155 & 115     \\
    EVEXTRA   & 8534 & 7056 & 5461 & 1103 & 902 & 700 & 2482 & 2077 & 1590  \\     
    \bottomrule
  \end{tabular}}}
  \caption{Statistics for the four datasets and their splits: the number of sentences (\#Sent), the number of sentences with triggers (\#Tr), and the number of relations after post-processing of MinIE triple extractions (\#Re).}
  \label{tab:dataset_stats}
\end{table}

\paragraph{MAVEN.} 

The MAssive eVENt detection dataset \citep{wang-etal-2020-maven} from the English Wikipedia domain is the largest freely available dataset suitable for TD. It covers more than 150 events. The size and coverage of event types make MAVEN an ideal source dataset for the domain transfer of TD. MAVEN comes with tokenized sentences and a predefined train, validation, and test split. However, since no gold test set labels were published, we use the official validation set as a test set (only to measure the source model performance on it) and randomly sample $20\%$ of sentences from the training data as a new validation set.

\paragraph{ACE 2005.} 

The Automatic Content Extraction dataset  \citep{doddington-etal-2004-automatic} is a widely used event detection dataset consisting predominantly of articles from various news sources in multiple languages. 
We use only the English train, validation, and test split, obtained with the standard ACE preprocessing tool,\footnote{\footnotesize\url{https://bit.ly/ace2005-preprocessing}} which we also use to obtain sentences and tokens. Although ACE is a sizable dataset, as noted by \citet{wang-etal-2020-maven}, many ACE sentences do not contain any triggers (cf.~Table~\ref{tab:dataset_stats}). 

\paragraph{EDNYT.} The event detection dataset of \citet{maisonnave2022detecting} was compiled from the New York Times articles on financial crises, which makes the dataset more topically focused than the other datasets. The dataset was not tokenized, but it came with a train-test split, with the test set comprising $10\%$ of the data. We obtain a validation set by randomly sampling $5\%$ of the train data. We use spaCy \citep{spacy} to tokenize the sentences. We discarded $3\%$ of sentences with trigger spans that could not be aligned with spaCy tokenization.

\paragraph{EVEXTRA.} The EVEXTRA dataset \citep{glavaš_šnajder_2015} is an English newspaper corpus annotated with event triggers. 
It comes tokenized but with no predefined split. We randomly assign sentences to train, validation, and test sets in a 70/10/20 ratio, respectively, ensuring that sentences from the same article end up in the same set. Less than $1\%$ of sentences were dropped because aligning the trigger annotations with tokens was impossible.



\paragraph{Relation Extraction.}

We use the rule-based OIE system MinIE \citep{gashteovski-etal-2017-minie} to extract subject-relation-object triples from sentences. MinIE has proven useful for many downstream tasks by the BenchIE benchmark and evaluation framework \citep{gashteovski-etal-2022-benchie}. However, it extracts all possible triples from the input text and introduces minor extraction errors, so we use a set of heuristics to post-process the results and improve the alignment of extracted relations and labeled triggers. To verify the alignment, we conduct a $\chi^2$ test of dependence on train sets of both source and target datasets, considering whether the same token is labeled as a relation and as a trigger. The dependence between variables was significant for all datasets ($p<.01$). A detailed description is given in Appendix~\ref{subsec:re_details}. First, we remove implicit\footnote{\footnotesize{OIE systems often incorporate binding tokens (like the copula \emph{is}), which do not have to be present in the text.}} triple extractions and discard all non-consecutive subject, relation, or object extractions. Further, we remove non-triples, relations with more than five tokens, and extractions not in the subject-relation-object order. Finally, we remove subject and object extraction information from the sentences and drop duplicates, leaving us only with relation extractions. Table~\ref{tab:dataset_stats} shows the final number of sentences containing relations in the post-processed datasets.


\subsection{Training Regimes}

In addition to using OIE relations with multi-task models to couple triggers with relations, we take inspiration from recent findings in language transfer \citep{meftah-etal-2021-hidden, schmidt-etal-2022-dont} and experiment with three transfer training regimes: \emph{joint training}, \emph{joint transfer}, and \emph{sequential transfer}. For the sake of completeness, we also consider \emph{in-domain training}, which reduces to fine-tuning each model on few-shot target domain examples.

\paragraph{Joint Training.}

The \emph{joint training} regime relies on mixed batches, adopted from the work on language transfer \citep{schmidt-etal-2022-dont}. A mixed batch consists predominantly of source trigger examples combined with a much lower fixed share of few-shot target trigger examples. Intuitively, having fewer few-shot examples should contribute to the update of model parameters with equal weight as the abundant source examples and ultimately prevent the model from overfitting on source data. We create mixed mini-batches consisting of $B \!=\! n \!+\! m$ examples, where $n$ are source examples, $m$ are randomly sampled few-shot target examples, and $n\!\gg\!m$. If more than $m$ few-shot examples are available, $m$ are consistently sampled from the few-shot pool. We fix $B\!=\!32$ with $n\!=\!27,m\!=\!5$ in our experiments. Fine-tuning is performed for a fixed number of epochs based on mixed mini-batch loss, calculated as the average of the source loss and $m$-shot target loss. In our experiments, \emph{joint training} amounts to mixed batch fine-tuning from either single- (TD) or multi-task (TD+RE) PLMs.


\paragraph{Joint Transfer.}

Similar to \emph{joint training}, the \emph{joint transfer} regime also uses mixed batches. However, instead of fine-tuning from PLM, we first train each PLM on source training data and then fine-tune with mixed batches in the same manner as in \emph{joint training}. \emph{Joint transfer} applied to multi-task models utilizes source OIE relations twice and target relations once during mixed batch fine-tuning.


\paragraph{Sequential Transfer.}

Analogously to \emph{joint transfer}, in the \emph{sequential transfer} regime, we fine-tune for a fixed number of epochs from the PLM trained on the source domain training data. However, unlike in \emph{joint transfer}, fine-tuning is done only with target few-shot examples.

\begin{table*}[!htb]
\small{\begin{tabular}{l|lccccccccc}
\toprule
\multicolumn{2}{c}{\multirow{2}{*}{\textbf{Training Regime}}} & \multicolumn{3}{c}{\textbf{ACE 2005 (0.706)}} & \multicolumn{3}{c}{\textbf{EDNYT (0.702)}} & \multicolumn{3}{c}{\textbf{EVEXTRA (0.893)}} \\
\cmidrule(lr){3-5} \cmidrule(lr){6-8} \cmidrule(lr){9-11}
\multicolumn{1}{l}{} & & {} \textbf{Vanilla} & \textbf{Implicit} & \textbf{Explicit} & \textbf{Vanilla} & \textbf{Implicit} & \textbf{Explicit} & \textbf{Vanilla} & \textbf{Implicit} & \textbf{Explicit} \\
\midrule
\multicolumn{1}{l}{} &   0-Shot          & 0.234           & 0.237          & \textbf{0.240} & 0.392          & 0.399          & \textbf{0.408} & 0.650           & 0.650          & \textbf{0.653} \\
\midrule
\multirow{6}{*}{\rotatebox[origin=c]{90}{\shortstack{joint \\ training}}} & 5-Shot        & 0.246           & 0.250          & \textbf{0.256} & 0.451         & 0.455          & \textbf{0.457} & 0.643           & 0.643          & \textbf{0.654} \\
{}  &   10-Shot    & 0.251           & 0.253          & \textbf{0.262} & 0.482          & \textbf{0.484} & \textbf{0.484} & 0.645           & 0.645          & \textbf{0.658} \\
{}  &   50-Shot       & 0.265           & 0.268          & \textbf{0.283} & 0.566          & \textbf{0.575} & 0.567          & 0.679           & 0.681          & \textbf{0.687} \\
{}  &   100-Shot      & 0.286           & 0.286          & \textbf{0.310} & 0.597          & \textbf{0.602} & 0.596          & 0.715           & 0.721          & \textbf{0.725} \\
{}  &   250-Shot      & 0.332           & 0.330          & \textbf{0.357} & 0.628          & \textbf{0.629} & \textbf{0.629} & 0.766           & \textbf{0.767} & 0.765          \\
{}  &   500-shot      & 0.382           & 0.378          & \textbf{0.398} & \textbf{0.649} & \textbf{0.649} & 0.646          & 0.793           & \textbf{0.798} & 0.792          \\
\midrule
\multirow{6}{*}{\rotatebox[origin=c]{90}{\shortstack{joint \\ transfer}}} &   5-Shot   & 0.248           & 0.248          & \textbf{0.254} & 0.433          & 0.436          & \textbf{0.440} & 0.631           & 0.633          & \textbf{0.636} \\
{}  &   10-Shot  & 0.251           & 0.250          & \textbf{0.256} & 0.448          & \textbf{0.451} & 0.450          & 0.632           & 0.634          & \textbf{0.638} \\
{}  &   50-Shot  & 0.262           & 0.265          & \textbf{0.267} & 0.524          & \textbf{0.536} & 0.507          & 0.650           & \textbf{0.656} & 0.648          \\
{} &    100-Shot    & 0.283           & 0.283          & \textbf{0.284} & 0.569          & \textbf{0.573} & 0.551          & 0.676           & \textbf{0.684} & 0.667          \\
{}  &   250-Shot & \textbf{0.328}           & \textbf{0.328} & 0.318          & 0.608          & \textbf{0.611} & 0.592          & 0.727           & \textbf{0.735} & 0.705          \\
{}  &   500-Shot & \textbf{0.388}  & 0.381          & 0.369          & 0.637          & \textbf{0.641} & 0.621          & 0.770           & \textbf{0.777} & 0.744          \\
\midrule
\multirow{6}{*}{\rotatebox[origin=c]{90}{\shortstack{sequential \\ transfer}}}   &   5-Shot     & \textbf{0.294}  & \textbf{0.294} & 0.276          & 0.458          & \textbf{0.466} & 0.448          & 0.659           & \textbf{0.661} & 0.653          \\
{}  &   10-Shot    & 0.372           & \textbf{0.374} & 0.330          & 0.512          & \textbf{0.521} & 0.490          & 0.688           & \textbf{0.693} & 0.680          \\
{}  &   50-Shot    & \textbf{0.511}  & 0.506          & 0.463          & 0.581          & \textbf{0.592} & 0.568          & 0.750           & \textbf{0.764} & 0.741          \\
{}  &   100-Shot   & 0.538           & \textbf{0.548} & 0.501          & 0.605          & \textbf{0.616} & 0.584          & 0.786           & \textbf{0.795} & 0.773          \\
{}  &   250-Shot   & \textbf{0.587}  & 0.577          & 0.556          & 0.631          & \textbf{0.644} & 0.607          & 0.824           & \textbf{0.835} & 0.813          \\
{}  &   500-Shot   & \textbf{0.610}  & 0.609          & 0.586          & \textbf{0.653} & 0.652          & 0.640          & 0.852           & \textbf{0.857} & 0.836          \\
\midrule
\midrule
\multirow{6}{*}{\rotatebox[origin=c]{90}{\shortstack{in-domain \\ training}}}  &   5-Shot              & 0.000           & 0.000          & 0.000          & 0.000          & 0.000          & 0.000          & 0.000           & 0.000          & 0.000          \\
{}  &   10-Shot              & 0.000           & 0.000          & 0.000          & 0.000          & 0.000          & 0.000          & 0.000           & 0.000          & 0.000          \\
{}  &   50-Shot              & 0.464           & \textbf{0.466} & 0.417          & \textbf{0.607} & 0.601          & 0.597          & 0.768           & \textbf{0.774} & 0.757          \\
{}  &   100-Shot             & 0.510           & \textbf{0.529} & 0.511          & 0.626          & \textbf{0.632} & 0.611          & 0.807           & \textbf{0.812} & 0.801          \\
{}  &   250-shot             & \textbf{0.570}  & 0.569          & 0.550          & 0.649          & \textbf{0.654} & 0.642          & 0.845           & \textbf{0.847} & 0.835          \\
{}  &    500-Shot            & 0.598           & \textbf{0.600} & 0.584          & 0.660          & 0.658          & \textbf{0.666} & 0.858           & \textbf{0.862} & 0.854          \\
\bottomrule
\end{tabular}}
\caption{TD domain transfer micro F1 scores when transferring from MAVEN as a source to ACE 2005, EDNYT, and EVEXTRA as targets (zero-shot, three few-shot transfer training regimes, and in-domain, with six varying numbers of shots). The numbers in parentheses next to the target dataset are the in-domain performance test set scores when using all target training data. \emph{Joint/in-domain training} -- target fine-tuning from PLM. \emph{Joint/sequential transfer} -- target fine-tuning from PLM trained for TD on MAVEN source training data. The best results by dataset and model per training regime are in \textbf{bold}. \emph{Implicit} and \emph{explicit} models leverage MinIE relation labels, unlike the \emph{vanilla} model. All reported results are averages of three runs. We report standard deviations in Appendix~\ref{subsec:add_results}.} 
\label{tab:main_results}
\end{table*}

\subsection{Training Details and Hyperparameters}

We briefly describe the training details (see Appendix~\ref{subsec:training_details} for more details). We use the RoBERTa-base \citep{liu2019roberta} PLM for token classification, implemented in \emph{Hugging Face} \citep{wolf-etal-2020-transformers}. We evaluate TD by micro F1 score on IOB2 tag predictions using strict matching, where the predicted output span must exactly match the expected output span. The models are trained with cross-entropy loss and Adam optimizer \citep{kingma2014adam} with the learning rate of $0.00001$ for $10$ epochs.

When training on the source domain, we use the source validation set to select the best model based on the TD micro F1 score. Specifically, we choose the model from the epoch that yields the highest TD validation performance.\footnote{\footnotesize We also experimented with selecting the model based on the MLM perplexity on the target validation set, but that led to worse performance than optimizing for TD F1 on the source validation set. The two options present a trade-off between learning TD adequately or adjusting to the target domain at the expense of TD performance.} 
Fine-tuning in \emph{joint/sequential transfer} regimes starts from the best model selected on the source validation set. In \emph{joint transfer} with the \emph{implicit} model, we perform mixed batch fine-tuning by averaging the source TD and target few-shot TD losses. Similarly, we average the source TD and RE losses with the target few-shot TD and RE losses in the \emph{joint transfer} with the \emph{explicit} model. 
Throughout experiments, we use a batch size of $B\!=\!32$. Also, we employ gradient clipping of model parameters to a maximum of $1.0$ before each mini-batch update. We do transfer experiments with $0, 5, 10, 50, 100, 250$, and $500$ shots. For MLM and \emph{in-domain training}, we update the models' parameters in an alternate fashion inside each epoch: first, based on target training data MLM loss, and then based on target few-shot loss. The MLM \emph{sequential transfer} is similar as without MLM. The difference is in the starting model, which is obtained by first training in the same described alternate fashion but with updates based on MLM loss on target training data and TD loss on source training data.

\section{Results and Discussion}

Table~\ref{tab:main_results} shows the main results of our experiments, with MinIE as a relation extractor for the multi-task models. \emph{Vanilla} is the sequence labeling PLM fine-tuned only for event TD, i.e., PLM with softmax token classifier on top trained on labeled event trigger spans. This model is trained in the same fashion as our proposed \emph{implicit} and \emph{explicit} variants, but without incorporating in any way the OIE relation information. For all experiments in this section, we average results over three seeds and report micro F1 TD scores on the held-out target test sets. For few-shot experiments, we additionally perform averaging on five different randomly sampled subsets from the target data training set. Moreover, we take precautions to ensure that samples from each draw are consistent across experiments and exclusively contain examples with triggers.

\subsection{Main Results} 

Zero-shot domain transfer of TD from MAVEN as the source to news datasets as targets exhibits noticeable negative transfer. The drops are massive compared to the performance of the models trained on all ACE 2005, EDNYT, or EVEXTRA training data. Even in this worst-case zero-shot setup, multi-task \emph{implicit} and \emph{explicit} models bring gains compared to \emph{vanilla} ones. Some interesting trends emerge when the number of shots increases. On average, relations help achieve higher target domain TD performance for a low-to-moderate number of shots. However, when the number of shots reaches $500$ (or even $250$ in some cases) target examples, the effects of relations become negligible, except for the EVEXTRA dataset, where the gains from relations are consistent regardless of the number of shots or training regime. When considering all training regimes, the \emph{implicit} model outperforms the \emph{explicit} model. Contrary to the findings from language transfer \citep{schmidt-etal-2022-dont}, \emph{joint} transfer training regimes were almost consistently worse compared to \emph{sequential transfer} and \emph{in-domain training}. 
These findings are of practical interest since \emph{joint} is worse performance-wise and takes far more resources and time to train. With $500$ shots, \emph{sequential transfer} and \emph{in-domain training} come close to the full in-domain training performance for each news dataset. For a low number of shots ($5$ and $10$), doing \emph{in-domain training} is useless, and in this case, \emph{sequential transfer} is a better option. However, a higher number of shots in combination with \emph{in-domain training} can lead to a better performance than \emph{sequential transfer}.

\subsection{Adding Auxiliary MLM Objective} 

\begin{figure*}[!htb]
\begin{center}
\begin{subfigure}{0.32\textwidth}
    \centering
    \includegraphics[width=1.0\linewidth]{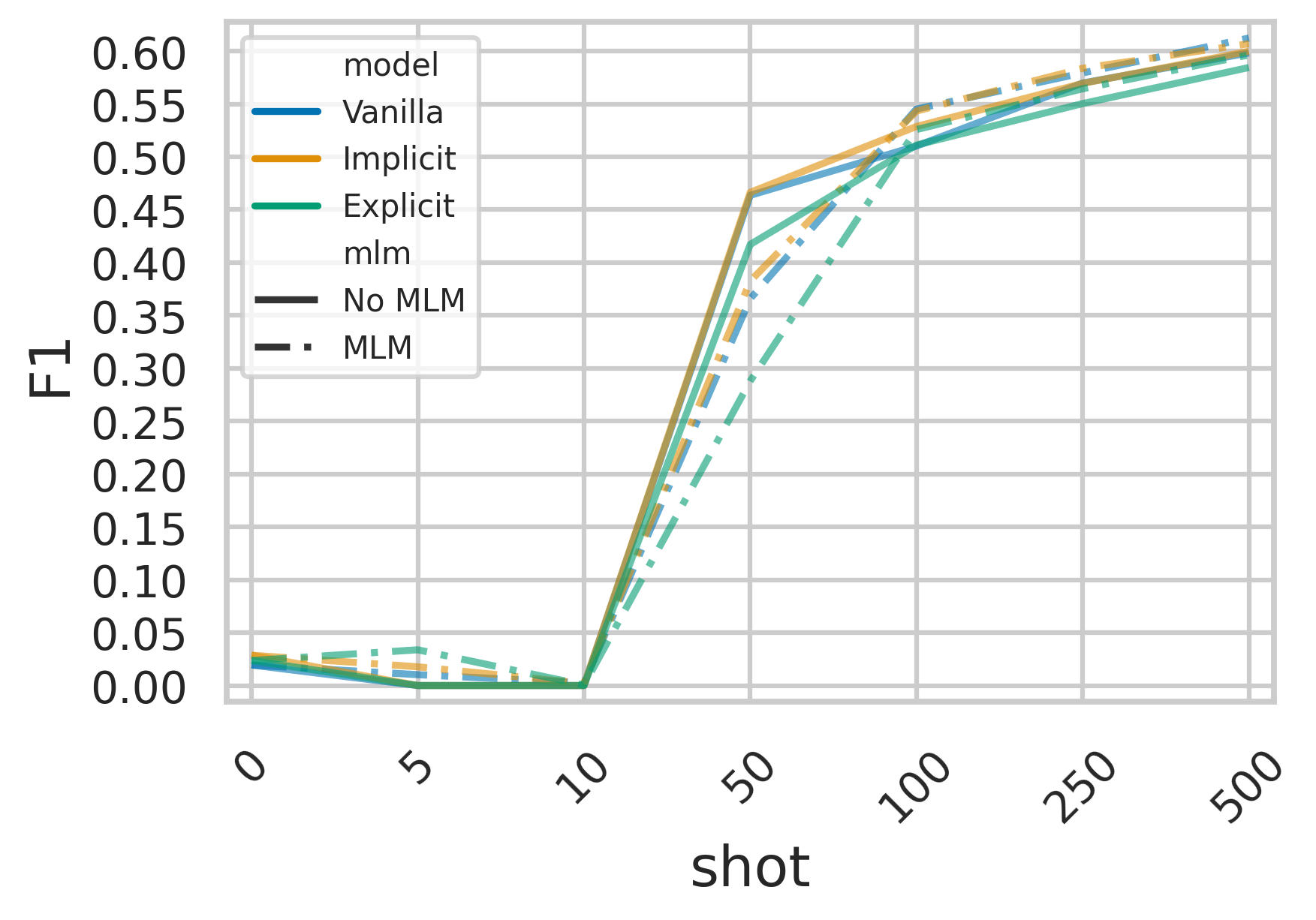} \\
    \includegraphics[width=1.0\linewidth]{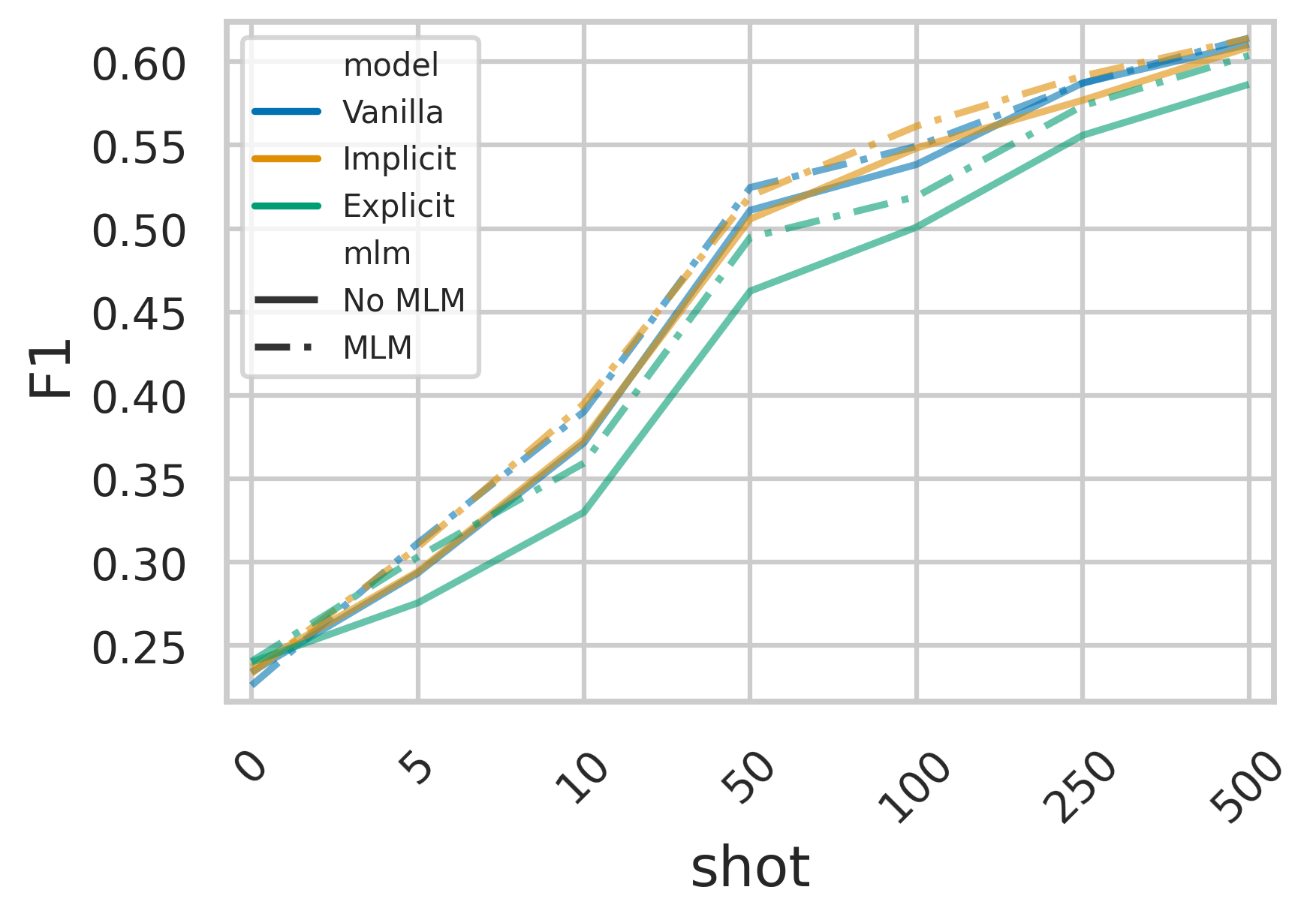}
    \caption{ACE 2005 (0.706)}
\end{subfigure}
\begin{subfigure}{0.32\textwidth}
    \centering
    \includegraphics[width=1.0\linewidth]{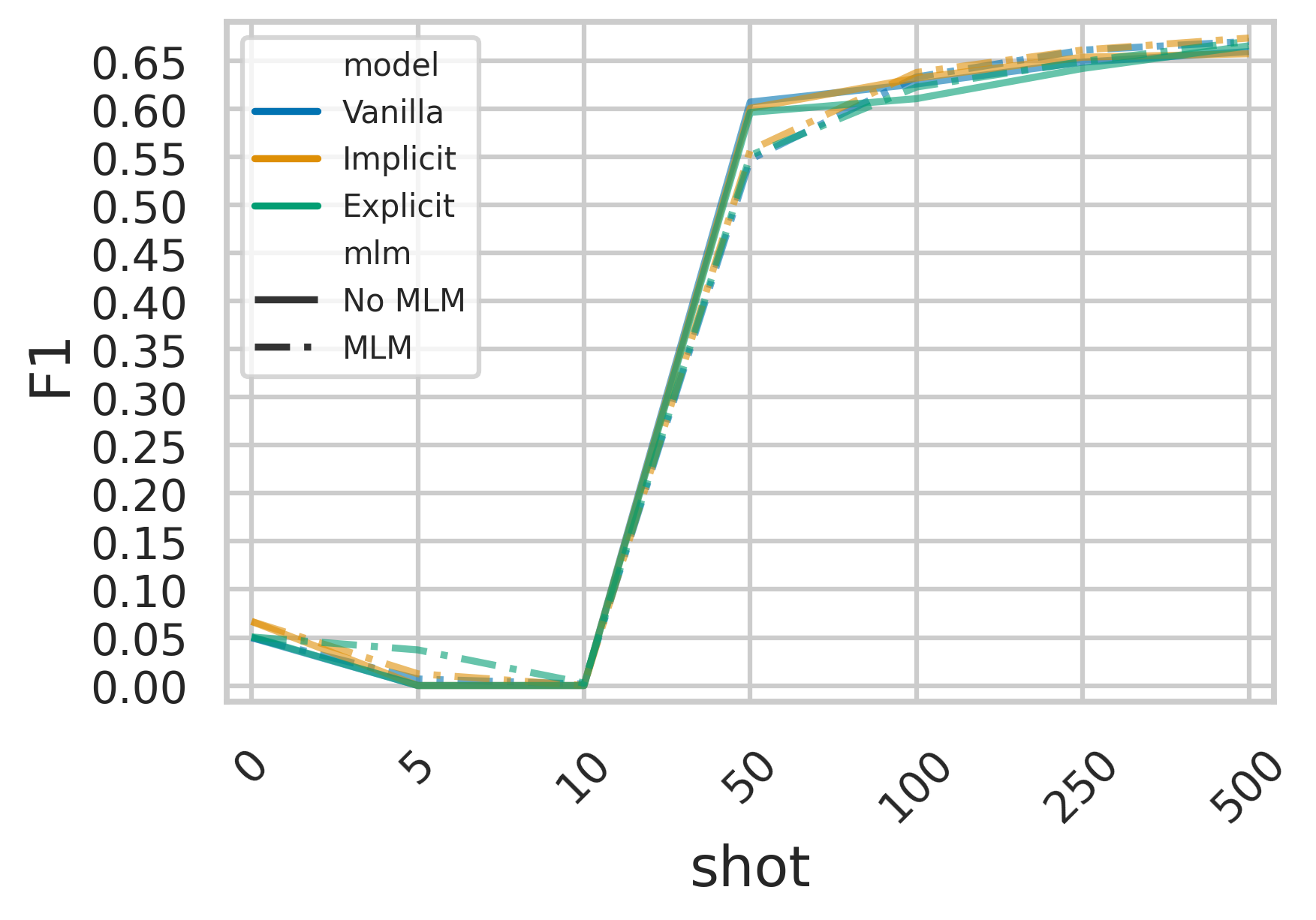} \\
    \includegraphics[width=1.0\linewidth]{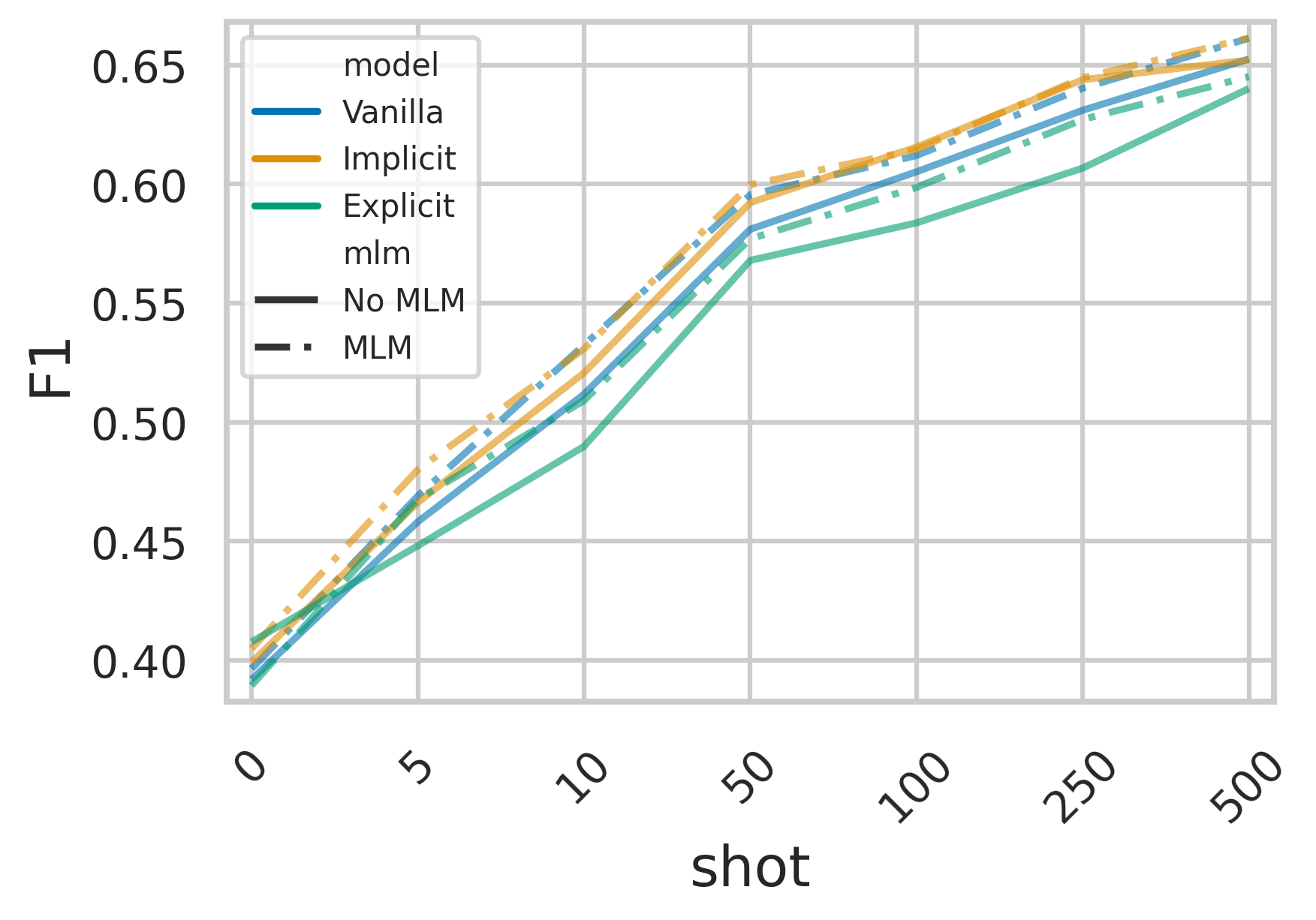}
    \caption{EDNYT (0.702)}
\end{subfigure}
\begin{subfigure}{0.32\textwidth}
    \centering
    \includegraphics[width=1.0\linewidth]{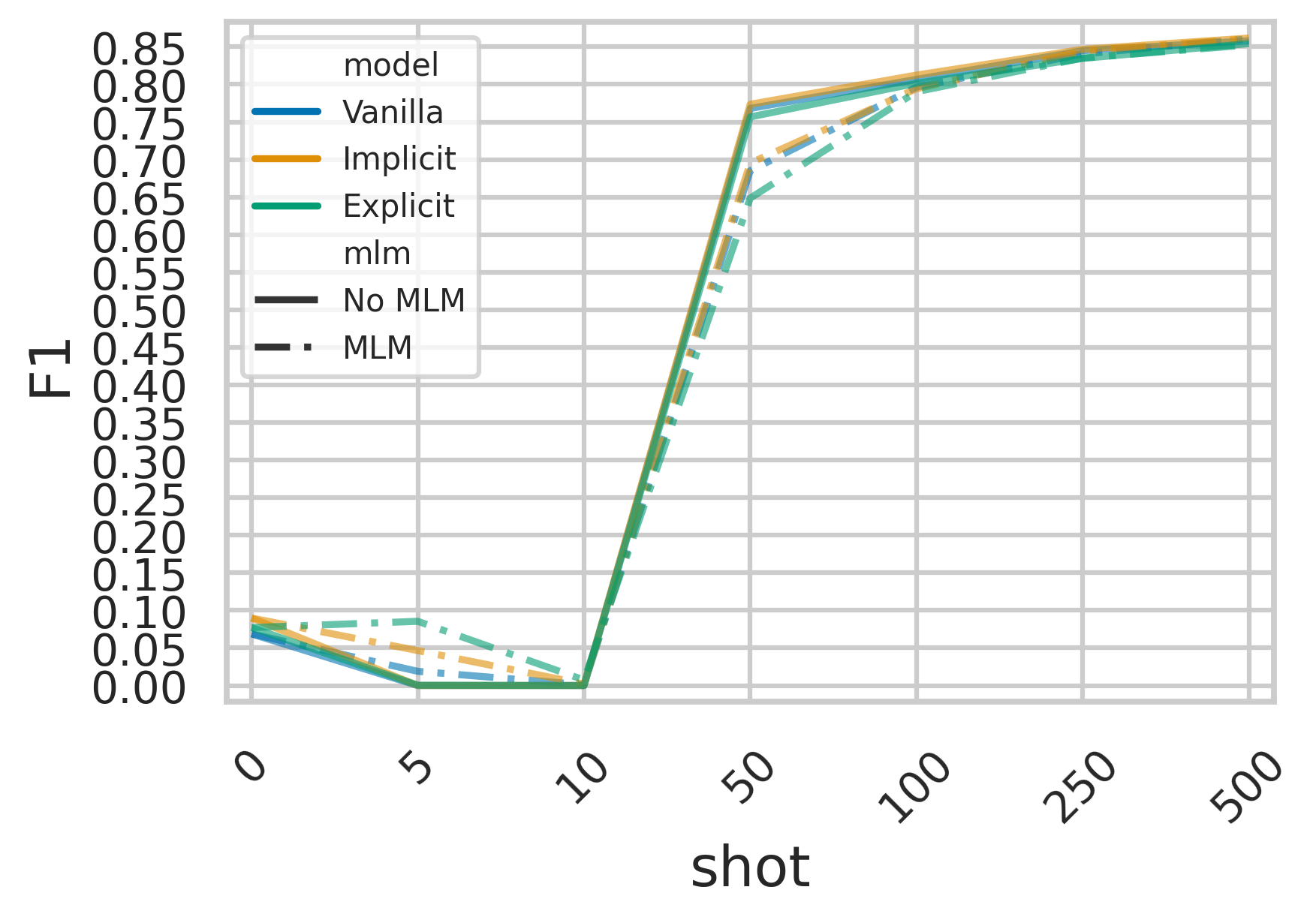} \\
    \includegraphics[width=1.0\linewidth]{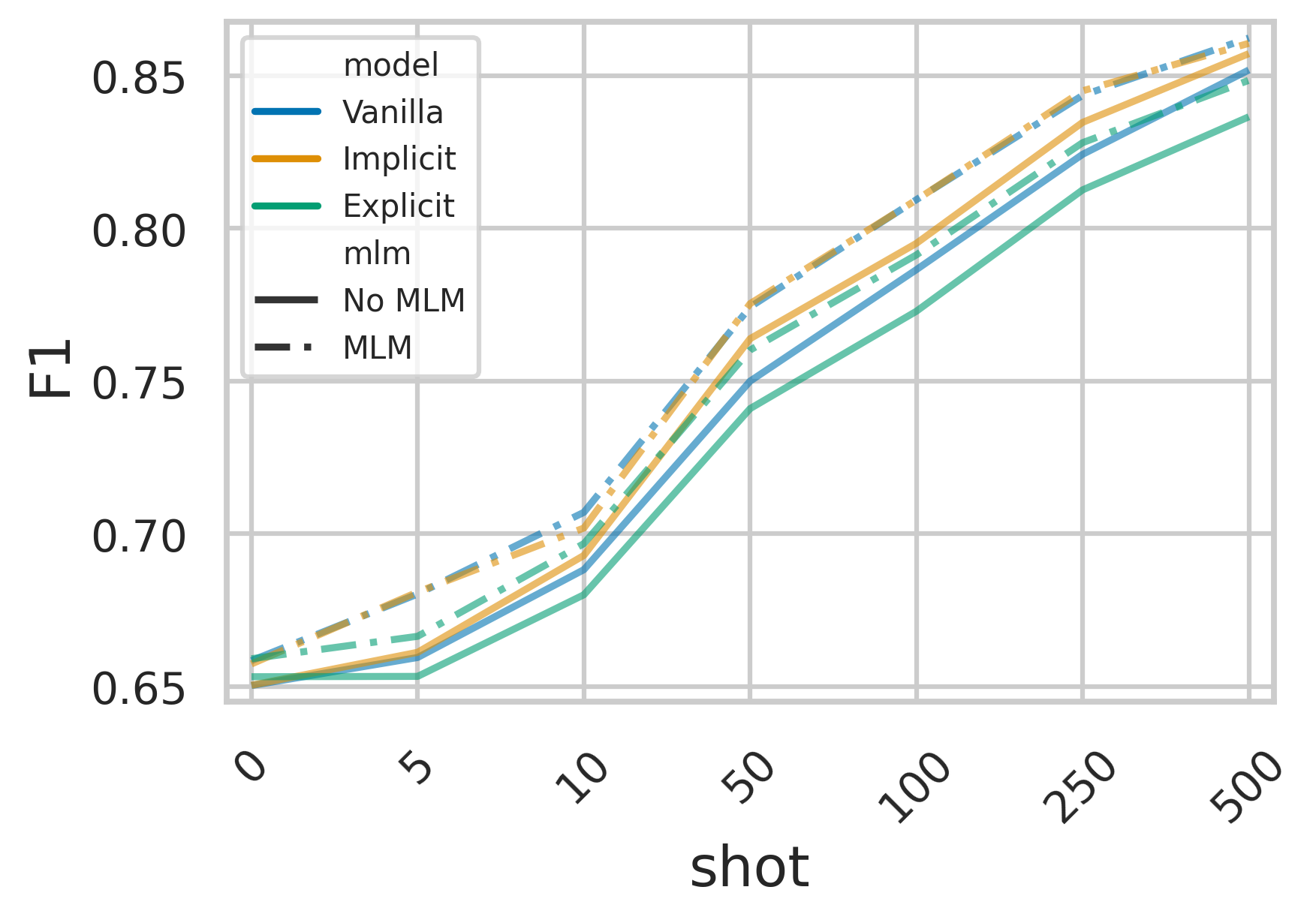}
    \caption{EVEXTRA (0.893)}
\end{subfigure}
\caption{TD domain transfer micro F1 scores when transferring from MAVEN as a source to ACE 2005, EDNYT, and EVEXTRA as targets (zero-shot, \emph{in-domain training}, and \emph{sequential transfer}, with six varying numbers of shots). The numbers in parentheses next to the target dataset are the in-domain performance test set scores when using all target training data. The upper three plots show \emph{in-domain training} results -- target fine-tuning starting from PLM. The lower three plots show \emph{sequential transfer} results -- target fine-tuning starting from PLM trained for TD on MAVEN source training data. Dash-dotted lines correspond to models with an auxiliary MLM objective on target domain training data. The x-axis shows the number of shots on an ordinal scale. \emph{Implicit} and \emph{explicit} models leverage MinIE relation labels, unlike the \emph{vanilla} model. All reported results are averages of three runs. The corresponding results in tabular form with standard deviations are in Appendix~\ref{subsec:add_results}.} 
\label{fig:mlm}
\end{center}
\end{figure*}

Building on recent findings from work on PLM domain adaptation \citep{gururangan-etal-2020-dont}, we investigate whether MLM can further boost TD transfer from Wikipedia to the news domain. Since \emph{joint} regimes were consistently worse in main results, we examine the MLM effect only for \emph{in-domain training} and \emph{sequential transfer}. We achieve this by adding token-level MLM as an auxiliary training objective through an extra MLM head in all model variants. The head's parameters are updated during training and not used during inference. Figure~\ref{fig:mlm} gives the results. 
\emph{Sequential transfer} proved to be more efficient than \emph{in-domain training}. On average, MLM with relations embodied into \emph{implicit} model in \emph{sequential transfer} regime outperforms the best results without MLM. An exception is the EVEXTRA dataset, where using OIE relations in conjunction with MLM and \emph{sequential transfer} does not lead to performance improvements compared to using only MLM.

\subsection{The Choice of the OIE System}

Finally, to examine if our results are specific to the OIE system, we replace MinIE with Stanford OIE. We post-process the relations in the same manner as for MinIE (cf.~Section~\ref{sec:exp_setup}). The experiments are conducted without MLM and for \emph{sequential transfer} and \emph{in-domain training} regimes. Table~\ref{tab:stanford_mini} shows the results. The difference between using MinIE and Stanford OIE is negligible for \emph{implicit} model but exists for \emph{explicit} model. Since \emph{explicit} outperformed \emph{implicit} in only five out of $156$ cases from Table~\ref{tab:stanford_mini}, we conclude that the gains from leveraging OIE relations in multi-task models are not due to the higher quality of MinIE extractions and persist for Stanford OIE. One can achieve similar, if not almost identical, gains using either extractor.

\begin{table*}[!htb]
\centering
\adjustbox{width=\textwidth}{\small{\begin{tabular}{l|lcccccccccccc}
\toprule
\multicolumn{2}{c}{\multirow{4}{*}{\textbf{Training Regime}}} & \multicolumn{4}{c}{\textbf{ACE 2005 (0.706)}} & \multicolumn{4}{c}{\textbf{EDNYT (0.702)}} & \multicolumn{4}{c}{\textbf{EVEXTRA (0.893)}} \\
\cmidrule(lr){3-6} \cmidrule(lr){7-10} \cmidrule(lr){11-14}
\multicolumn{1}{l}{} & {} & \multicolumn{2}{c}{\textbf{MinIE}} & \multicolumn{2}{c}{\textbf{Stanford OIE}} & \multicolumn{2}{c}{\textbf{MinIE}} & \multicolumn{2}{c}{\textbf{Stanford OIE}} & \multicolumn{2}{c}{\textbf{MinIE}} & \multicolumn{2}{c}{\textbf{Stanford OIE}} \\
\cmidrule(lr){3-4} \cmidrule(lr){5-6} \cmidrule(lr){7-8} \cmidrule(lr){9-10} \cmidrule(lr){11-12} \cmidrule(lr){13-14}
\multicolumn{1}{l}{} & {} & \textbf{Implicit} & \textbf{Explicit} & \textbf{Implicit} & \textbf{Explicit} & \textbf{Implicit} & \textbf{Explicit} & \textbf{Implicit} & \textbf{Explicit} & \textbf{Implicit} & \textbf{Explicit} & \textbf{Implicit} & \textbf{Explicit} \\
\midrule
\multicolumn{1}{l}{} & 0-Shot   & 0.237    &0.240    & 0.237    & \textbf{0.242}    & 0.399    & \textbf{0.408}    & 0.401    & 0.406    &0.650    & 0.653    &0.650    & \textbf{0.657}     \\
\midrule
\multirow{6}{*}{\rotatebox[origin=c]{90}{\shortstack{sequential \\ transfer}}}   & 5-Shot   & 0.294    & 0.276    & \textbf{0.296}    & 0.283    & 0.466    & 0.448    & \textbf{0.468}    & 0.464    & \textbf{0.661}    & 0.653    & \textbf{0.661}    & 0.658     \\
{} & 10-Shot  & 0.374    & 0.330    & \textbf{0.375}    &0.350    & \textbf{0.521}    & 0.490    &0.520    & 0.512    & \textbf{0.693}    &0.680    & \textbf{0.693}    & 0.688     \\
{} & 50-Shot  & \textbf{0.506}    & 0.463    & \textbf{0.506}    & 0.476    & \textbf{0.592}    & 0.568    & 0.591    &0.570    & \textbf{0.764}    & 0.741    & 0.763    & 0.747     \\
{} & 100-Shot & \textbf{0.548}    & 0.501    & \textbf{0.548}    & 0.525    & \textbf{0.616}    & 0.584    & 0.615    & 0.587    & 0.795    & 0.773    & \textbf{0.796}    & 0.775     \\
{} & 250-Shot & \textbf{0.577}    & 0.556    & \textbf{0.577}    & 0.568    & 0.644    & 0.607    & \textbf{0.647}    & 0.602    & \textbf{0.835}    & 0.813    & 0.834    & 0.818     \\
{} & 500-Shot & \textbf{0.609}    & 0.586    & 0.602    & 0.584    & 0.652    &0.640    & \textbf{0.653}    & 0.627    & \textbf{0.857}    & 0.836    & 0.856    & 0.845 \\
\midrule
\midrule
\multirow{6}{*}{\rotatebox[origin=c]{90}{\shortstack{in-domain \\ training}}} & 5-Shot   &0.000       &0.000       &0.000       &0.000       &0.000       &0.000       &0.000       &0.000       &0.000       &0.000       &0.000       &0.000        \\
{} & 10-Shot  &0.000       &0.000       &0.000       &0.000       &0.000       &0.000       &0.000       &0.000       &0.000       &0.000       &0.000       &0.000        \\
{} & 50-Shot  & 0.466    & 0.417    & \textbf{0.467}    & 0.446    & 0.601    & 0.597    & 0.601    & \textbf{0.605}    & 0.774    & 0.757    & \textbf{0.775}    & 0.765     \\
{} & 100-Shot & \textbf{0.529}    & 0.511    & \textbf{0.529}    & 0.515    & 0.632    & 0.611    & \textbf{0.633}    & 0.615    & 0.812    & 0.801    & \textbf{0.814}    & 0.805     \\
{} & 250-Shot & \textbf{0.569}    &0.550    & \textbf{0.569}    & 0.557    & \textbf{0.654}    & 0.642    & 0.652    & 0.638    & \textbf{0.847}    & 0.835    & 0.846    & 0.840     \\
{} & 500-Shot & \textbf{0.600}      & 0.584    & 0.598    & 0.585    & 0.658    & \textbf{0.666}    & 0.657    & 0.662    & \textbf{0.862}    & 0.854    & 0.861    & 0.852     \\
\bottomrule
\end{tabular}}}
\caption{TD domain transfer micro F1 scores when transferring from MAVEN as a source to ACE 2005, EDNYT, and EVEXTRA as targets w.r.t. MinIE and Stanford OIE systems (zero-shot, \emph{sequential transfer}, and \emph{in-domain training}, with six varying numbers of shots). The numbers in parentheses next to the target dataset are the in-domain performance test set scores when using all target training data. \emph{Sequential transfer} -- target fine-tuning from PLM trained for TD on MAVEN source training data. \emph{In-domain training} -- target fine-tuning from PLM. The best results by dataset, \emph{implicit} or \emph{explicit} relation-leveraging models, per training regime and OIE system, are in \textbf{bold}. All reported results are averages of three runs.} 
\label{tab:stanford_mini}
\end{table*}

\section{Conclusion}

We showed that OIE relations can be utilized to improve the domain transfer of trigger detection (TD) in zero- and few-shot setups. The best improvements were achieved with \emph{implicit} multi-task model and \emph{sequential transfer} training regime. We also demonstrated that more substantial gains can be reached when combining OIE relations with MLM as an auxiliary task. This is especially evident for the models pre-trained with TD task on the source domain and with MLM training objective on the target domain in the \emph{implicit} multi-task model. Replacing MinIE with Stanford OIE revealed that gains on the target domain for the TD task persist when using the other OIE extractor. 

Future work may further explore the potential of OIE for improving domain transfer of TD on diverse datasets and domains, such as the cybersecurity \citep{man-duc-trong-etal-2020-introducing}, literature \citep{sims-etal-2019-literary}, and biomedical \citep{kim-etal-2009-overview} domains.
Applying the coupling concept to other NLP tasks, such as event argument detection or named entity recognition, where OIE extractions might enhance the in- and out-of-domain performance, is another exciting future work direction.

\section{Limitations}

Our experiments were limited by the available computing resources. For reliability, in our experiments, we report performance scores averaged over three runs (differing in random seeds). Similarly, we sampled the few-shot examples five times. Averaging over larger samples would make the results even more reliable. Furthermore, the results of few-shot experiments can sometimes turn out to be misleading due to the high variance of the sample of examples. Fixing the learning rate and some other hyperparameters across experiments may have resulted in suboptimal adaptation to the trigger detection task in both source and target domains. Moreover, all experiments were done only with RoBERTa-base; using a different suitable PLM might yield further insights. Finally, our experiments were limited to datasets in the English language; further insights may be gained by extending to cross-lingual trigger detection domain transfer, more transfer directions, and datasets.


\section{Ethical Considerations}

Developing models for automated event detection comes with inherent risks, including the potential for misuse and unintended consequences. The ability to autonomously extract events from sensitive data raises possible ethical concerns, especially in the context of enhanced domain transfer. Combining open information systems with trigger detection models for improved domain transfer reduces the effort of event extraction from sensitive data in a novel domain when only a handful of annotated examples from that domain can be obtained.


\bibliography{anthology,custom}

\appendix

\section{Appendix} \label{sec:appendix}

\subsection{Relation Extraction Details} \label{subsec:re_details}

During the relation extraction with the OIE system, implicit triples and long relations can appear. We filter out both implicit triples and long relations (longer than five tokens) as it has been shown that these relations are noisy \citep{broscheit-etal-2020-predict-new}, and implicit relations cannot be used for token classification since they introduce tokens that are not present in the text. For example, if the OIE system is presented with the sentence: ``President Biden right now stands really worried about future economic growth.'' it might extract (i) implicit triple (``Biden''; ``is''; ``President'') and (ii) triple with long relation (``President Biden''; ``right now stands really worried about''; ``future economic growth''). Our heuristics would drop both extractions, and the implicit extraction would also be filtered out on account of not being in the order subject-relation-object in the input sentence. Also, we filter out all extractions that are incomplete triples, i.e., are missing either subject, relation, or object. If, after that, there are still multiple relation extractions for the same sentence, we try to merge the remaining relations. The merging process is designed to keep all the relations if the tokens are not shared between them. In the case of shared tokens, we keep only the relation extraction with the highest number of tokens that make up the relation. Finally, subject and object extractions are dropped, only the relations are kept, and if our heuristics filter out all the relation extractions for the sentence, we do not discard it but consider it a sentence without relations and use it for training as an example with all ``outside'' token labels based on IOB2 tagging scheme. We apply the OIE system, and this described post-processing, to each split of the source and target datasets.\footnote{\footnotesize{Relation extractor is always shared between domains.}}

\subsection{Experimental Setup Details} \label{subsec:training_details}

\paragraph{Training.} The total GPU usage for all the experiments amounts to $1280$ hours on \emph{Ampere A100} GPU. We use the RoBERTa-base model with $125$ million parameters. The input sequences are not lowercased. Since RoBERTa-base works on input split into subwords, the TD cross-entropy loss is adjusted to take into account only the first token of each tokenized word from the input sequence. Our preliminary experiments found incorporating a learning rate scheduler is beneficial. We use a multiplicative learning rate scheduler with a multiplying factor of $0.99$, which multiplies the learning rate in each epoch, lowering it throughout training. For each mini-batch, padding is applied to match the length of the longest example in the batch. 

\paragraph{Hyperparameter Optimization.} When training on the source domain, the \emph{implicit} model is additionally optimized on the source validation set (based on the TD micro F1 score) with a simple grid search over the dimension of the trainable OIE-label embeddings $d$ and the learning rate for it. We try dimensions of $10, 50, 100$, and $300$ and learning rates of $0.0001, 0.00005$, and $0.00001$. When performing target few-shot fine-tuning in \emph{joint transfer} and \emph{sequential transfer}, we fix the dimension to the one that produced the highest source validation set TD micro F1 score. In the \emph{joint training} and \emph{in-domain training} experiments, we arbitrarily fix the embedding size of the \emph{implicit} model to $300$ and $10$ across all the experiments, respectively.

\paragraph{Auxiliary MLM Objective.} We use a token-level masking probability of $15\%$, and the masking procedure is inherited from \citet{devlin-etal-2019-bert}. Specifically, out of $15\%$ of randomly chosen tokens, we mask $80\%$ tokens, replace $10\%$ tokens with random tokens from the vocabulary, and leave the remaining $10\%$ of tokens unchanged.

\subsection{Additional Results} \label{subsec:add_results}

\begin{table*}[!htb]
\begin{subtable}{\textwidth}
\small{\begin{tabular}{l|lccccccccc}
\toprule
\multicolumn{2}{c}{\multirow{2}{*}{\textbf{Training Regime}}} & \multicolumn{3}{c}{\textbf{ACE 2005 (0.706)}} & \multicolumn{3}{c}{\textbf{EDNYT (0.702)}} & \multicolumn{3}{c}{\textbf{EVEXTRA (0.893)}} \\
\cmidrule(lr){3-5} \cmidrule(lr){6-8} \cmidrule(lr){9-11}
\multicolumn{1}{l}{} & & {} \textbf{Vanilla} & \textbf{Implicit} & \textbf{Explicit} & \textbf{Vanilla} & \textbf{Implicit} & \textbf{Explicit} & \textbf{Vanilla} & \textbf{Implicit} & \textbf{Explicit} \\
\midrule
\multicolumn{1}{l}{} &   0-Shot          & 0.234           & 0.237          & \textbf{0.240} & 0.392          & 0.399          & \textbf{0.408} & 0.650           & 0.650          & \textbf{0.653} \\
\midrule
\multirow{6}{*}{\rotatebox[origin=c]{90}{\shortstack{sequential \\ transfer}}}   &   5-Shot     & \textbf{0.294}  & \textbf{0.294} & 0.276          & 0.458          & \textbf{0.466} & 0.448          & 0.659           & \textbf{0.661} & 0.653          \\
{}  &   10-Shot    & 0.372           & \textbf{0.374} & 0.330          & 0.512          & \textbf{0.521} & 0.490          & 0.688           & \textbf{0.693} & 0.680          \\
{}  &   50-Shot    & \textbf{0.511}  & 0.506          & 0.463          & 0.581          & \textbf{0.592} & 0.568          & 0.750           & \textbf{0.764} & 0.741          \\
{}  &   100-Shot   & 0.538           & \textbf{0.548} & 0.501          & 0.605          & \textbf{0.616} & 0.584          & 0.786           & \textbf{0.795} & 0.773          \\
{}  &   250-Shot   & \textbf{0.587}  & 0.577          & 0.556          & 0.631          & \textbf{0.644} & 0.607          & 0.824           & \textbf{0.835} & 0.813          \\
{}  &   500-Shot   & \textbf{0.610}  & 0.609          & 0.586          & \textbf{0.653} & 0.652          & 0.640          & 0.852           & \textbf{0.857} & 0.836          \\
\midrule
\midrule
\multirow{6}{*}{\rotatebox[origin=c]{90}{\shortstack{in-domain \\ training}}}  &   5-Shot              & 0.000           & 0.000          & 0.000          & 0.000          & 0.000          & 0.000          & 0.000           & 0.000          & 0.000          \\
{}  &   10-Shot              & 0.000           & 0.000          & 0.000          & 0.000          & 0.000          & 0.000          & 0.000           & 0.000          & 0.000          \\
{}  &   50-Shot              & 0.464           & \textbf{0.466} & 0.417          & \textbf{0.607} & 0.601          & 0.597          & 0.768           & \textbf{0.774} & 0.757          \\
{}  &   100-Shot             & 0.510           & \textbf{0.529} & 0.511          & 0.626          & \textbf{0.632} & 0.611          & 0.807           & \textbf{0.812} & 0.801          \\
{}  &   250-shot             & \textbf{0.570}  & 0.569          & 0.550          & 0.649          & \textbf{0.654} & 0.642          & 0.845           & \textbf{0.847} & 0.835          \\
{}  &    500-Shot            & 0.598           & \textbf{0.600} & 0.584          & 0.660          & 0.658          & \textbf{0.666} & 0.858           & \textbf{0.862} & 0.854          \\
\bottomrule
\end{tabular}}
\subcaption{Without MLM.}
\end{subtable}

\vspace{0.15cm}

\begin{subtable}{\linewidth}
\small{\begin{tabular}{l|lccccccccc}
\toprule
\multicolumn{2}{c}{\multirow{2}{*}{\textbf{Training Regime}}} & \multicolumn{3}{c}{\textbf{ACE 2005 (0.706)}} & \multicolumn{3}{c}{\textbf{EDNYT (0.702)}} & \multicolumn{3}{c}{\textbf{EVEXTRA (0.893)}} \\
\cmidrule(lr){3-5} \cmidrule(lr){6-8} \cmidrule(lr){9-11}
\multicolumn{1}{l}{} & & {} \textbf{Vanilla} & \textbf{Implicit} & \textbf{Explicit} & \textbf{Vanilla} & \textbf{Implicit} & \textbf{Explicit} & \textbf{Vanilla} & \textbf{Implicit} & \textbf{Explicit} \\
\midrule
\multicolumn{1}{l}{} &   0-Shot          & 0.226 &              0.233 &              \textbf{0.241} & 0.396 &              \textbf{0.405} &              0.389 & 0.658 &              0.657 &              \textbf{0.659} \\
\midrule
\multirow{6}{*}{\rotatebox[origin=c]{90}{\shortstack{sequential \\ transfer}}}   &   5-Shot     & \textbf{0.311} &              0.309 &              0.303          & 0.469 &              \textbf{0.480} &              0.468          & 0.680 &              \textbf{0.681} &              0.666          \\
{}  &   10-Shot    & 0.390 &              \textbf{0.395} &              0.359          & \textbf{0.532} &              0.531 &              0.509          & \textbf{0.707} &              0.702 &              0.697          \\
{}  &   50-Shot    & \textbf{0.525} &              0.520 &              0.495          & 0.595 &              \textbf{0.600} &              0.577          & 0.774 &              \textbf{0.775} &              0.760          \\
{}  &   100-Shot   & 0.549 &              \textbf{0.561} &              0.519          & 0.612 &              \textbf{0.615} &              0.599          & \textbf{0.809} &              \textbf{0.809} &              0.791          \\
{}  &   250-Shot   & 0.587 &              \textbf{0.591} &              0.574          & 0.640 &              \textbf{0.645} &              0.627           & 0.843 &              \textbf{0.845} &              0.828          \\
{}  &   500-Shot   & \textbf{0.614} &              \textbf{0.614} &              0.604          & \textbf{0.661} &              \textbf{0.661} &              0.645          & \textbf{0.862} &              0.861 &              0.848          \\
\midrule
\midrule
\multirow{6}{*}{\rotatebox[origin=c]{90}{\shortstack{in-domain \\ training}}}  &   5-Shot              & 0.010 &              0.018 &              \textbf{0.034}          & 0.007 &              0.012 &              \textbf{0.037}          & 0.019 &              0.046 &              \textbf{0.085}          \\
{}  &   10-Shot              & \textbf{0.002} &              \textbf{0.002} &              0.000          & 0.002 &              0.000 &              \textbf{0.003}          & 0.001 &              0.002 &              \textbf{0.007}          \\
{}  &   50-Shot              & 0.366 &              \textbf{0.383} &              0.288          & 0.548 &              \textbf{0.557} &              0.552          & 0.685 &              \textbf{0.695} &              0.649          \\
{}  &   100-Shot             & \textbf{0.545} &              0.543 &              0.526          & 0.633 &              \textbf{0.638} &              0.623          & \textbf{0.796} &              0.794 &              0.790          \\
{}  &   250-shot             & 0.579 &              \textbf{0.584} &              0.564          & \textbf{0.661} &              \textbf{0.661} &              0.650          & 0.841 &              \textbf{0.844} &              0.835          \\
{}  &    500-Shot            & \textbf{0.612} &              0.607 &              0.596          & 0.670 &              \textbf{0.674} &              0.671 & \textbf{0.861} &              \textbf{0.861} &              0.852          \\
\bottomrule
\end{tabular}}
\subcaption{With MLM.}
\end{subtable}

\caption{TD domain transfer micro F1 scores when transferring from MAVEN as a source to ACE 2005, EDNYT, and EVEXTRA as targets (zero-shot, \emph{sequential transfer}, and \emph{in-domain training}, with six varying numbers of shots). The numbers in parentheses next to the target dataset are the in-domain performance scores when using all target training data. \emph{In-domain training} results -- target fine-tuning starting from PLM. \emph{Sequential transfer} results -- target fine-tuning starting from PLM trained for TD on MAVEN source training data. Table (a) shows results without an auxiliary MLM objective, while Table (b) depicts results with an auxiliary MLM training objective on target domain training data. \emph{Implicit} and \emph{explicit} models leverage MinIE relation labels, unlike the \emph{vanilla} model. All reported results are averages of three runs.} 
\label{tab:mlm_results}
\end{table*}

\begin{table*}[!htb]
\small{\begin{tabular}{l|lccccccccc}
\toprule
\multicolumn{2}{c}{\multirow{2}{*}{\textbf{Training Regime}}} & \multicolumn{3}{c}{\textbf{ACE 2005 (0.706)}} & \multicolumn{3}{c}{\textbf{EDNYT (0.702)}} & \multicolumn{3}{c}{\textbf{EVEXTRA (0.893)}} \\
\cmidrule(lr){3-5} \cmidrule(lr){6-8} \cmidrule(lr){9-11}
\multicolumn{1}{l}{} & & {} \textbf{Vanilla} & \textbf{Implicit} & \textbf{Explicit} & \textbf{Vanilla} & \textbf{Implicit} & \textbf{Explicit} & \textbf{Vanilla} & \textbf{Implicit} & \textbf{Explicit} \\
\midrule
\multicolumn{1}{l}{} &   

0-Shot          & 0.005 &              0.003 &              0.003  & 0.007 &              0.009 &              0.005  & 0.003 &              0.002 &              0.004  \\
\midrule
\multirow{6}{*}{\rotatebox[origin=c]{90}{\shortstack{joint \\ training}}} & 

5-Shot        & 0.008 &              0.001 &              0.003  & 0.014 &              0.015 &              0.011  & 0.004 &              0.001 &              0.002  \\
{}  &   

10-Shot    & 0.003 &              0.003 &              0.006  & 0.014 &              0.010 &              0.010  & 0.003 &              0.002 &              0.005  \\
{}  &   

50-Shot       & 0.006 &              0.005 &              0.011  & 0.018 &              0.009 &              0.012  & 0.005 &              0.008 &              0.007  \\
{}  &   

100-Shot      & 0.005 &              0.002 &              0.010  & 0.011 &              0.003 &              0.004  & 0.008 &              0.005 &              0.008  \\
{}  &   

250-Shot      & 0.009 &              0.003 &              0.010  & 0.010 &              0.004 &              0.007  & 0.010 &              0.008 &              0.004  \\
{}  &   

500-shot      & 0.013 &              0.010 &              0.006  & 0.009 &              0.001 &              0.006  & 0.008 &              0.002 &              0.005  \\
\midrule
\multirow{6}{*}{\rotatebox[origin=c]{90}{\shortstack{joint \\ transfer}}} &   

5-Shot   & 0.010 &              0.005 &              0.004  & 0.018 &              0.012 &              0.018  & 0.004 &              0.006 &              0.002  \\
{}  &   

10-Shot  & 0.011 &              0.006 &              0.005  & 0.014 &              0.009 &              0.018  & 0.004 &              0.006 &              0.003  \\
{}  &   

50-Shot  & 0.007 &              0.007 &              0.005  & 0.005 &              0.002 &              0.006  & 0.001 &              0.005 &              0.004  \\
{} &    

100-Shot    & 0.005 &              0.006 &              0.005  & 0.005 &              0.003 &              0.014  & 0.004 &              0.004 &              0.005  \\
{}  &   

250-Shot & 0.012 &              0.007 &              0.014  & 0.009 &              0.015 &              0.009  & 0.005 &              0.001 &              0.011  \\
{}  &   

500-Shot & 0.008 &              0.008 &              0.021  & 0.009 &              0.006 &              0.008  & 0.006 &              0.005 &              0.004  \\
\midrule
\multirow{6}{*}{\rotatebox[origin=c]{90}{\shortstack{sequential \\ transfer}}}   &   

5-Shot     & 0.014 &              0.016 &              0.014  & 0.022 &              0.024 &              0.025  & 0.012 &              0.003 &              0.003  \\
{}  &   

10-Shot    & 0.012 &              0.016 &              0.020  & 0.013 &              0.013 &              0.017  & 0.011 &              0.005 &              0.003  \\'
{}  &  

50-Shot    & 0.011 &              0.006 &              0.003  & 0.004 &              0.010 &              0.006  & 0.011 &              0.010 &              0.003  \\
{}  &   

100-Shot   & 0.003 &              0.015 &              0.013  & 0.003 &              0.012 &              0.004  & 0.009 &              0.008 &              0.002  \\
{}  &   

250-Shot   & 0.007 &              0.006 &              0.012  & 0.004 &              0.012 &              0.013  & 0.009 &              0.005 &              0.005  \\
{}  &  

500-Shot   & 0.004 &              0.010 &              0.002  & 0.004 &              0.009 &              0.002  & 0.004 &              0.003 &              0.005  \\
\midrule
\midrule
\multirow{6}{*}{\rotatebox[origin=c]{90}{\shortstack{in-domain \\ training}}}  &   

5-Shot              & 0.000 &              0.000 &              0.000  & 0.000 &              0.000 &              0.000  & 0.000 &              0.000 &              0.001  \\
{}  &   

10-Shot              & 0.000 &              0.000 &              0.000  & 0.000 &              0.000 &              0.000  & 0.000 &              0.000 &              0.001  \\
{}  &   

50-Shot              & 0.013 &              0.013 &              0.034  & 0.007 &              0.003 &              0.010  & 0.009 &              0.012 &              0.009  \\
{}  &   

100-Shot             & 0.009 &              0.006 &              0.012  & 0.001 &              0.008 &              0.010  & 0.004 &              0.004 &              0.007  \\
{}  &   

250-shot             & 0.001 &              0.004 &              0.017  & 0.012 &              0.010 &              0.007  & 0.003 &              0.005 &              0.006  \\
{}  &    

500-Shot            & 0.008 &              0.004 &              0.006  & 0.004 &              0.010 &              0.010  & 0.003 &              0.006 &              0.003  \\
\bottomrule
\end{tabular}}
\caption{Standard deviation of TD domain transfer micro F1 scores from Table~\ref{tab:main_results}. All reported results are averages of three runs.}
\label{tab:main_results_stdev}
\end{table*}

\begin{table*}[!htb]
\begin{subtable}{\textwidth}
\small{\begin{tabular}{l|lccccccccc}
\toprule
\multicolumn{2}{c}{\multirow{2}{*}{\textbf{Training Regime}}} & \multicolumn{3}{c}{\textbf{ACE 2005 (0.706)}} & \multicolumn{3}{c}{\textbf{EDNYT (0.702)}} & \multicolumn{3}{c}{\textbf{EVEXTRA (0.893)}} \\
\cmidrule(lr){3-5} \cmidrule(lr){6-8} \cmidrule(lr){9-11}
\multicolumn{1}{l}{} & & {} \textbf{Vanilla} & \textbf{Implicit} & \textbf{Explicit} & \textbf{Vanilla} & \textbf{Implicit} & \textbf{Explicit} & \textbf{Vanilla} & \textbf{Implicit} & \textbf{Explicit} \\
\midrule

\multicolumn{1}{l}{} &   

0-Shot          & 0.005 &              0.003 &              0.003  & 0.007 &              0.009 &              0.005  & 0.003 &              0.002 &              0.004  \\

\midrule

\multirow{6}{*}{\rotatebox[origin=c]{90}{\shortstack{sequential \\ transfer}}}   &   

5-Shot     & 0.014 &              0.016 &              0.014  & 0.022 &              0.024 &              0.025  & 0.012 &              0.003 &              0.003  \\
{}  &   

10-Shot    & 0.012 &              0.016 &              0.020  & 0.013 &              0.013 &              0.017  & 0.011 &              0.005 &              0.003  \\'
{}  &  

50-Shot    & 0.011 &              0.006 &              0.003  & 0.004 &              0.010 &              0.006  & 0.011 &              0.010 &              0.003  \\
{}  &   

100-Shot   & 0.003 &              0.015 &              0.013  & 0.003 &              0.012 &              0.004  & 0.009 &              0.008 &              0.002  \\
{}  &   

250-Shot   & 0.007 &              0.006 &              0.012  & 0.004 &              0.012 &              0.013  & 0.009 &              0.005 &              0.005  \\
{}  &  

500-Shot   & 0.004 &              0.010 &              0.002  & 0.004 &              0.009 &              0.002  & 0.004 &              0.003 &              0.005  \\
\midrule
\midrule
\multirow{6}{*}{\rotatebox[origin=c]{90}{\shortstack{in-domain \\ training}}}  &   

5-Shot              & 0.000 &              0.000 &              0.000  & 0.000 &              0.000 &              0.000  & 0.000 &              0.000 &              0.001  \\
{}  &   

10-Shot              & 0.000 &              0.000 &              0.000  & 0.000 &              0.000 &              0.000  & 0.000 &              0.000 &              0.001  \\
{}  &   

50-Shot              & 0.013 &              0.013 &              0.034  & 0.007 &              0.003 &              0.010  & 0.009 &              0.012 &              0.009  \\
{}  &   

100-Shot             & 0.009 &              0.006 &              0.012  & 0.001 &              0.008 &              0.010  & 0.004 &              0.004 &              0.007  \\
{}  &   

250-shot             & 0.001 &              0.004 &              0.017  & 0.012 &              0.010 &              0.007  & 0.003 &              0.005 &              0.006  \\
{}  &    

500-Shot            & 0.008 &              0.004 &              0.006  & 0.004 &              0.010 &              0.010  & 0.003 &              0.006 &              0.003  \\
\bottomrule
\end{tabular}}
\subcaption{Without MLM.}
\end{subtable}

\vspace{0.15cm}

\begin{subtable}{\linewidth}
\small{\begin{tabular}{l|lccccccccc}
\toprule
\multicolumn{2}{c}{\multirow{2}{*}{\textbf{Training Regime}}} & \multicolumn{3}{c}{\textbf{ACE 2005 (0.706)}} & \multicolumn{3}{c}{\textbf{EDNYT (0.702)}} & \multicolumn{3}{c}{\textbf{EVEXTRA (0.893)}} \\
\cmidrule(lr){3-5} \cmidrule(lr){6-8} \cmidrule(lr){9-11}
\multicolumn{1}{l}{} & & {} \textbf{Vanilla} & \textbf{Implicit} & \textbf{Explicit} & \textbf{Vanilla} & \textbf{Implicit} & \textbf{Explicit} & \textbf{Vanilla} & \textbf{Implicit} & \textbf{Explicit} \\
\midrule
\multicolumn{1}{l}{} &   0-Shot          & 0.005 &              0.002 &              0.008  & 0.005 &              0.001 &              0.012  & 0.001 &              0.005 &              0.001  \\
\midrule
\multirow{6}{*}{\rotatebox[origin=c]{90}{\shortstack{sequential \\ transfer}}}   &   5-Shot     & 0.017 &              0.018 &              0.016  & 0.022 &              0.019 &              0.007  & 0.003 &              0.003 &              0.003  \\

{}  &   10-Shot    & 0.028 &              0.018 &              0.020  & 0.012 &              0.013 &              0.006  & 0.003 &              0.004 &              0.007  \\
{}  &   50-Shot    & 0.008 &              0.014 &              0.021  & 0.006 &              0.010 &              0.005  & 0.005 &              0.007 &              0.003  \\
{}  &   100-Shot   & 0.009 &              0.012 &              0.012  & 0.001 &              0.006 &              0.002  & 0.001 &              0.004 &              0.004  \\
{}  &   250-Shot   & 0.013 &              0.013 &              0.005  & 0.005 &              0.008 &              0.005  & 0.004 &              0.004 &              0.003  \\
{}  &   500-Shot   & 0.006 &              0.007 &              0.015  & 0.008 &              0.003 &              0.007  & 0.002 &              0.003 &              0.002  \\
\midrule
\midrule
\multirow{6}{*}{\rotatebox[origin=c]{90}{\shortstack{in-domain \\ training}}}  &   5-Shot              & 0.003 &              0.016 &              0.032  & 0.005 &              0.011 &              0.032  & 0.007 &              0.037 &              0.072  \\
{}  &   10-Shot              & 0.002 &              0.002 &              0.001  & 0.003 &              0.000 &              0.002  & 0.001 &              0.002 &              0.007  \\
{}  &   50-Shot              & 0.042 &              0.042 &              0.067  & 0.008 &              0.010 &              0.022  & 0.026 &              0.006 &              0.033  \\
{}  &   100-Shot             & 0.009 &              0.013 &              0.004  & 0.008 &              0.002 &              0.002  & 0.009 &              0.013 &              0.002  \\
{}  &   250-shot             & 0.008 &              0.002 &              0.004  & 0.001 &              0.004 &              0.002  & 0.004 &              0.002 &              0.003  \\
{}  &    500-Shot            & 0.018 &              0.008 &              0.014  & 0.002 &              0.002 &              0.001  & 0.001 &              0.002 &              0.002  \\
\bottomrule
\end{tabular}}
\subcaption{With MLM.}
\end{subtable}

\caption{Standard deviation of TD domain transfer micro F1 scores from Table~\ref{tab:mlm_results}. All reported results are averages of three runs.}
\label{tab:mlm_results_stdev}
\end{table*}

\end{document}